\documentclass[10pt,journal,compsoc]{IEEEtran}
\usepackage{times}
\usepackage{epsfig}
\usepackage{graphicx}
\usepackage{amsmath}
\usepackage{amssymb}
\usepackage{multirow}
\usepackage{array}
\usepackage{bm}
\usepackage{tabularx}
\usepackage{amsmath}
\usepackage{amssymb}
\usepackage{latexsym}
\usepackage{algpseudocode}
\usepackage{tabularx}
\usepackage{booktabs}
\usepackage{url}
\usepackage{svg}
\usepackage{bbding}
\usepackage{algorithmicx}
\usepackage{algorithm}
\usepackage{graphicx}
\usepackage{amssymb,subfigure,cases}
\usepackage{booktabs}
\usepackage{tabularx,booktabs}
\usepackage{times}
\usepackage{color}
\usepackage{multirow}
\usepackage{hyperref}

\usepackage{caption} %改变图表标题
\usepackage{booktabs} %调整表格线与上下内容的间隔
\usepackage{longtable}%调用跨页表格
\usepackage{multirow} %多行合并
\usepackage{array} %调用公式宏包的命令应放在调用定理宏包命令之前，也能控制表格

\ifCLASSOPTIONcompsoc
  \usepackage[nocompress]{cite}
\else
  \usepackage{cite}
\fi

\begin{document}

\title{Vehicle Perception from Satellite}

\author{Bin Zhao,~\IEEEmembership{Member, IEEE}, Pengfei Han, Xuelong Li,~\IEEEmembership{Fellow, IEEE}

}

\IEEEtitleabstractindextext{%
\begin{abstract}
	
Satellites are capable of capturing high-resolution videos. It makes vehicle perception from satellite become possible. Compared to street surveillance, drive recorder or other equipments, satellite videos provide a much broader city-scale view, so that the global dynamic scene of the traffic are captured and displayed. Traffic monitoring from satellite is a new task with great potential applications, including traffic jams prediction, path planning, vehicle dispatching, \emph{etc.}. Practically, limited by the resolution and view, the captured vehicles are very tiny (a few pixels) and move slowly. Worse still, these satellites are in Low Earth Orbit (LEO) to capture such high-resolution videos, so the background is also moving. Under this circumstance, traffic monitoring from the satellite view is an extremely challenging task. To attract more researchers into this field, we build a large-scale benchmark for traffic monitoring from satellite. It supports several tasks, including tiny object detection, counting and density estimation. The dataset is constructed based on 12 satellite videos and 14 synthetic videos recorded from GTA-V. They are separated into 408 video clips, which contain 7,336 real satellite images and 1,960 synthetic images. 128,801 vehicles are annotated totally, and the number of vehicles in each image varies from 0 to 101. Several classic and state-of-the-art approaches in traditional computer vision are evaluated on the datasets, so as to compare the performance of different approaches, analyze the challenges in this task, and discuss the future prospects. The dataset is available at: \href{https://github.com/Chenxi1510/Vehicle-Perception-from-Satellite-Videos}{https://github.com/Chenxi1510/Vehicle-Perception-from-Satellite-Videos}.
	
\end{abstract}

}

\maketitle
\IEEEdisplaynontitleabstractindextext
\IEEEpeerreviewmaketitle
\section{Introduction}

Recently, with the significant progress of aerospace technology, the commercial satellites are able to capture the Very High Resolution (VHR) videos~\cite{zhang2021moving,shi2022accurate}. For example, the Jilin-1 satellites can observe the earth with the spatial resolution of 0.72m \cite{gu2020semi}. SkySat-1 satellites provide the VHR videos with the resolution around 1m \cite{zhang2020error}. These satellites can monitor the ground dynamically in city-scale, where the vehicles can be seen clearly \cite{gao2023global}. VHR videos from satellites provide a new perspective for traffic monitoring.

Traffic monitoring from satellite is quite different from those street surveillance cameras on the ground. It has a variety of advantages:

1) Geographical Boundlessness: Satellite transcends terrestrial geographical constraints, encompassing urban, rural, and remote terrains. It orbits dynamically, affording a panoramic vantage point for comprehensive vehicular oversight. In contrast, surveillance cameras are constrained to urban centers and pivotal transport hubs. The installation of surveillance cameras worldwide is prohibitively costly due to extreme geographical conditions. Yet, such coverage is pivotal for traffic monitoring in remote areas and holistic network analysis.

2) Real-time and Periodicity: Satellites furnish nearly real-time imagery and data, facilitating immediate traffic monitoring and emergency response. Furthermore, routine satellite missions yield long-term time-series data for traffic trend and pattern analysis. They swiftly monitor traffic conditions during natural disasters or emergencies, providing crucial information for rescue operations.

3) Wide-area Surveillance: Satellite traffic monitoring spans vast geographical regions, from cities to broader territories. This has immense value for urban planning, traffic management, and natural disaster monitoring, offering a global perspective. In contrast to ground-based surveillance cameras that can only observe traffic in intersections, streets, or roundabouts, satellites can monitor sprawling areas spanning square kilometers. Their expansive field of view enables the surveillance of city-scale traffic, providing a fresh perspective for traffic control, analysis, and planning.

4) Monitoring Diverse Transport Modes: Satellites can monitor not only road traffic but also railways, aviation, and maritime transport, fostering multimodal traffic research. Additionally, satellites can complement surveillance cameras, enabling layered traffic monitoring in cities. Surveillance cameras offer localized and detailed insights, while satellites provide global traffic monitoring. They can be integrated to monitor the traffic from the ground view and sky view hierarchically.

\begin{figure}[tp]
	\centering
	\includegraphics[width=0.48\textwidth]{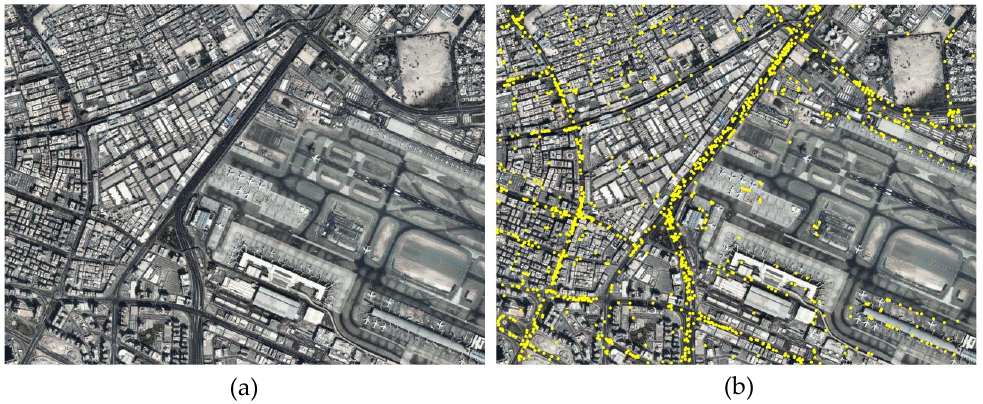}
	\caption{Examples of (a) satellite images captured over Dubai International Airport, and (b) the 1,214 vehicles in the scene highlighted by yellow points.}\label{Fig1}
\end{figure}

Apart from the above advantages, traffic monitoring from satellite is also a challenging task:

1) The vehicles in satellite videos are tiny. Limited by the resolution, the vehicles only contain a few pixels (10 pixels or so) and lack of appearance information. Some of them cannot even be recognized unless they are moving, as depicted in Fig. 1. The detection and counting of such tiny vehicles are very difficult. Traditional object detection and counting approaches cannot deal with them effectively and efficiently.

2) The movement in satellite videos is very complicated. As depicted in Fig. 2 (a) and (b), the background and vehicles are both moving. Practically, the movement of background is correlated to the view of satellite and the height of buildings, since the video only captures the 2D projection of the 3D movement. As a result, it is hard to separate the moving vehicles from the background. Moreover, the gradual movement of the satellite engenders localized misalignment and dynamic intensity variations in some stationary background objects. These changes serve no purpose in discerning the motion of objects and may potentially give rise to motion artifacts.

3) The satellite videos are full of noise. As evidenced in Fig. 2 (c)-(f). The frames captured are subject to an array of perturbations, including obstructions from edifices, clouds in the sky, shadows cast by sunlight, and mirrored reflections on ground-level glass surfaces. Under such circumstances, the local contrast between the background and targets diminishes significantly. Owing to the complexity of the background and the presence of image noise, targets occasionally amalgamate with the chaotic background, resulting in localized obscurity. This noise exerts substantial interference in the detection of vehicles.

4) The LEO satellites are incapable of perpetually hovering over a certain urban domain. In such instances, satellites are unable to sustain prolonged and continuous traffic surveillance. Additionally, when the satellite platform is fixated on a specific area of interest, the acquired satellite video evinces localized positional discrepancies and alterations in local intensity, attributable to immobile objects. These objects may be erroneously discerned as mobile entities, thereby exacerbating the incidence of false positives in the data.

\begin{figure}[tp]
	\centering
	\includegraphics[width=0.48\textwidth]{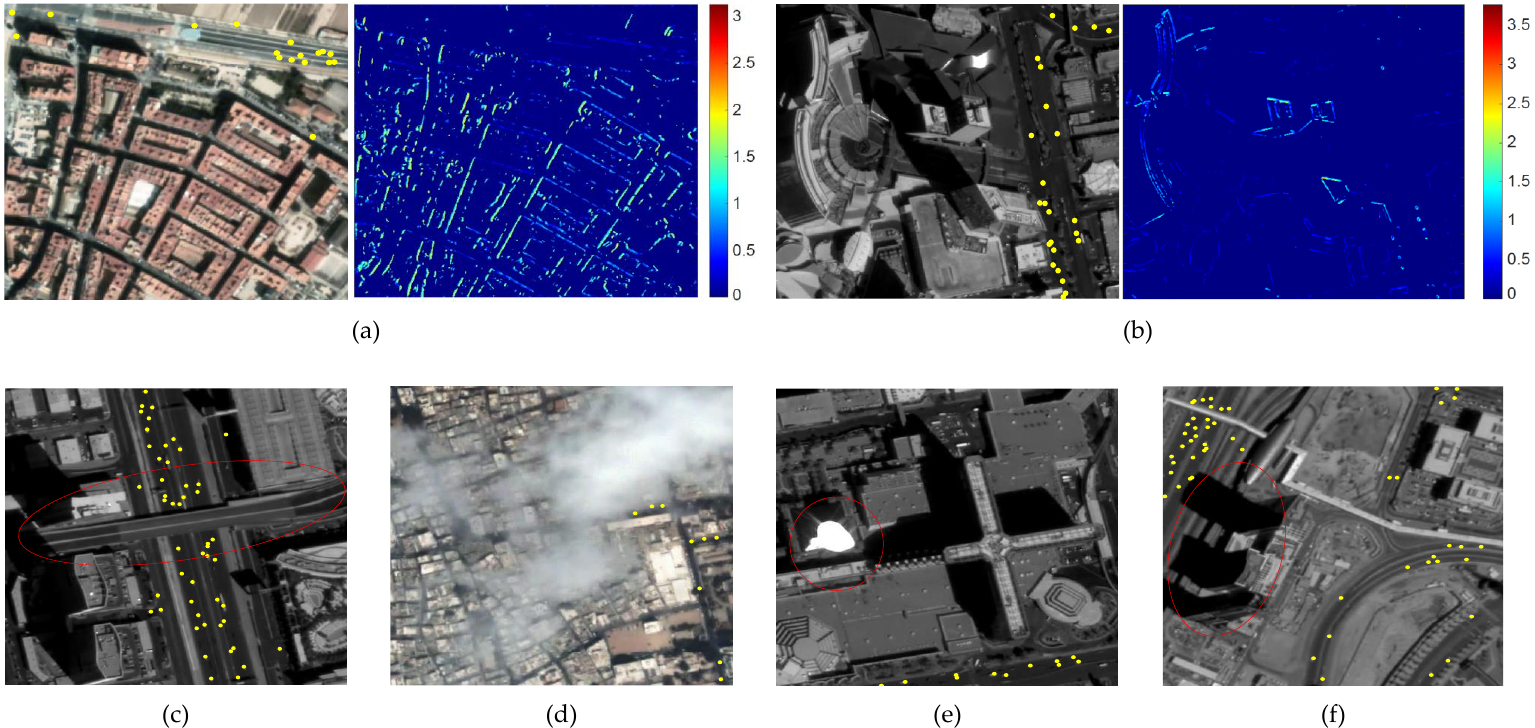}
	\caption{Challenges in vehicle perception from satellite. The optical flow maps in (a) and (b) indicate the movements are complex and uneven. (c), (d), (e) and (f) display the noise in satellite videos. They are shelter, clouds, specularity and shadow from left to right. }\label{Fig2}
\end{figure}

\subsection{Motivation}

Traffic monitoring from satellite is still in the preliminary stage. The lack of released satellite videos and large-scale annotated datasets are the key factors to limit the development. To the best of our knowledge, only two videos are released with partially annotations \cite{ao2019needles,zhang2021moving}. The lack of data leads to the following dilemmas in the research:

 1) Most of existing approaches just evaluate their performance on one or two videos. Such little test samples are not enough to verify the effectiveness. Worse still, some of the annotations or videos are not released publicly, which causes interference to the fair comparison of different approaches. 

 2) Deep learning approaches are not applicable to this field. Most existing approaches are developed based on traditional detectors, since the annotated samples are not enough for training. However, deep learning has become the mainstream and surpasses traditional approaches in most computer vision tasks. The performance of this field is impeded by the lack of annotated data.

\subsection{Overview}

Practically, the annotations of vehicles in satellite videos are difficult, since the frames are in city-scale, the movements of background are uneven, the sunlights are varying, and the vehicles are very tiny and cannot even be recognized by human eyes. In this paper, we make efforts to overcome these difficulties, and construct a large-scale dataset for traffic monitoring from satellite, which is named as TMS. TMS is composed of 408 videos collected from 12 real satellite videos and 14 synthetic videos of GTA-V, where the synthetic videos are utilized to make up for the lack of available satellite videos. In each video, the coordinates of vehicles are annotated at 1 frame per second. The numbers of vehicles in each frame vary from 0 to 101, and 128,801 vehicles are annotated totally. With the help of the TMS dataset, three tasks are developed for traffic monitoring from satellite, \emph{i.e.,} Tiny Object Detection (TOD), VEhicle Counting (VEC) and Traffic Density Estimation (TDE). Numbers of classic and state-of-the-art approaches are evaluated, including both traditional and deep learning approaches, where the challenges of each task are analyzed and the insights for researchers are presented.

\subsection{Contributions}

Overall, the main contributions of this paper can be summarized as follows:

1) The largest satellite video dataset is constructed for traffic monitoring. It can promote the research in this field by attracting deep learning approaches and provide an evaluation platform for different approaches.

2) The synthetic videos from GTA-V are integrated with real videos to further augment the scale of real dataset. They are annotated automatically, and provide a new perspective to relieve the problem of lacking real data. 

3) Three tasks are performed on this benchmark, including Tiny Object Detection (TOD), VEhicle Counting (VEC), and Traffic Density Estimation (TDE), so as to promote the development of traffic monitoring from satellite.

\section{Related Works}\label{section2}

In the following subsections, the related tasks in remote sensing, computer vision, and the recently proposed approaches of traffic monitoring from satellite are reviewed.

\subsection{Object Detection in Remote Sensing}
Object detection is a long standing task in remote sensing, which is also the basis of traffic monitoring from satellite. Numbers of datasets are constructed, \emph{e.g.}, TAS \cite{heitz2008learning}, SZTAKI-INRIA \cite{benedek2011building}, NWPU VHR-10 \cite{cheng2016learning}, HRSC2016 \cite{liu2016ship}, DOTA \cite{xia2018dota}, \emph{etc.}. The images are mainly collected from satellite, aerial plane, and other platforms, such as Google Earth, Tianditu and Quickbird. The annotated objects are in multiple categories. Vehicle is the most popular category for object detection, which shows its importance in remote sensing \cite{liu2015fast}. Earlier object detection approaches are developed based on template matching, geometry modeling, context knowledge, and low-level feature extraction \cite{felzenszwalb2010cascade}. However, object detection in remote sensing is quite complex with the factors of noise, size, lights and background \cite{zhang2021salient}. These traditional approaches are not robust enough to be generalized to different situations. Recently, deep learning is employed in this task under the support of large-scale datasets. Most of them are modified from mainstream object detection approaches in natural scene images, \emph{e.g.}, Faster RCNN \cite{DBLP:conf/nips/RenHGS15}, SSD \cite{DBLP:conf/eccv/LiuAESRFB16}, YOLO \cite{DBLP:conf/cvpr/RedmonDGF16}. By taking advantages of the non-linear learning ability, deep learning approaches surpass traditional approaches, and boost the performance significantly.

\subsection{Relevant Tasks in Computer Vision}

Object detection is a classic computer vision task for natural scene images. Traditional approaches follow a pipeline of region selection (\emph{e.g.}, superpixels \cite{yan2015object}, sliding window \cite{lampert2008beyond}, selective search \cite{uijlings2013selective}), feature extraction (\emph{e.g.}, SIFT \cite{zhao2012flip}, HOG \cite{dalal2005histograms}) and classifier (\emph{e.g.}, SVM \cite{malisiewicz2011ensemble}, Adaboost \cite{viola2001rapid}). However, traditional approaches and hand-crafted features are not capable enough to generalize to the variance of size, shape, occlusion and noise in object detection. Till now, deep learning approaches are taking the leading position, including RCNN series \cite{DBLP:conf/nips/RenHGS15,girshick2015fast}, YOLO series \cite{redmon2016you,redmon2017yolo9000,bochkovskiy2020yolov4}, \emph{etc.}. They are mostly developed based on challenges, \emph{i.e.}, PASCAL VOC\footnote{\url {http://host.robots.ox.ac.uk/pascal/VOC/index.html}}, ILSVRC\footnote{\url {https://image-net.org/challenges/LSVRC/2017/}} and MS-COCO Detection\footnote{\url {https://cocodataset.org/\#home}}. Object detection in natural images can inspire vehicle detection in satellite videos. Object counting and density estimation are important for crowd analysis as well as traffic congestion prediction. Apart from adopting object detectors directly, CNN-based regression models are widely used in this task. Most of them are developed based on Fully Convolutional Network (FCN) \cite{long2015fully} with the instance-level or image-level annotations. TRANCOS \cite{TRANCOSdataset_IbPRIA2015} and VisDrone2019 Veheicle \cite{zhu2020vision} are two datasets specially designed for vehicle counting and density estimation, which can benefit traffic monitoring from satellite.

\subsection{Traffic Monitoring from Satellite}

With the release of several satellite videos from Jilin-1 and SkySat-1, traffic monitoring from satellite draws increasing attention \cite{zhang2021moving,ao2019needles}. Most works are presented in recent years. In \cite{kopsiaftis2015vehicle}, the vehicles are detected by background subtraction, where mathematic morphology and statistical analysis are utilized to estimate background of each frame. The performance is evaluated on one satellite video from SkySat-1. The low-rank matrix decomposition is modified in \cite{zhang2020error}, in order to model background and foreground with the regularization of low-rank and sparsity. Furthermore, matrix decomposition assisted with moving confidence is developed in \cite{zhang2021moving}, which can promote the motion of vehicles meanwhile suppress that of the background. The performance is evaluated on two satellite videos and two surveillance videos. Overall, traffic monitoring from satellite is still in its early stages, and requires large-scale datasets and benchmarks to advance the development.

\section{The TMS Dataset}\label{section3}

\subsection{Data Collection and Preprocessing}

We tried our best to collect the satellite videos that can be used for traffic monitoring from satellite. The real part of TMS is composed of 12 full satellite videos. They are captured by the non-stationary satellite platform, \emph{i.e.}, Jilin-1 and SkySat-1, with the ground sample distance of 1m or so. In this case, the vehicles only contain 5-20 pixels. The videos are recorded over the sky of Boston, Dubai, Valencia, Jeddah, LasVegas, Hong Kong, Aleppo, Bangkok, and Tokyo. The captured scenes include city street, airport, suburbs, and port. Five of the videos are freely provided by the official website of Jilin-1\footnote{\url{https://mall.charmingglobe.com/Sampledata}}, and the others are crawled from YouTube\footnote{\url{https://www.youtube.com/}}, since the officially released versions are not available.

The synthetic part of TMS contains 14 videos obtained from the game of GTA-V. GTA-V builds a fictional city with Los Angels as the prototype, namely, Los Santos. The map in the game covers 252 square kilometers. The scene rendering, lights, shadow, weather effects and other conditions are quite similar to those in the real world, so that the players can immerse themselves into it. Furthermore, GTA-V is allowed to be developed by players for non-commercial use, such as academic use. Different from capturing videos from the egocentric view of the player, the videos are recorded from the satellite view of GTA-V in this paper. These videos display different city streets or suburbs in the map of the game. They are recorded by the screen recording software of Windows 10, \emph{i.e.}, Xbox Game Bar. 

The real and synthetic satellite videos are combined together to form the TMS dataset. Each video is with the resolution of 1080x1920 or 3072x4096, which covers an area of several square kilometers. For simplicity, these full satellite videos are segmented spatially into subareas. In this case, those videos with the resolution of 1080x1920 are segmented evenly into 540x480 subareas by 2x4 grid, and those videos with the resolution of 3072x4096 are segmented evenly into 512x512 subareas by 6x8 grid. Totally, 408 videos are obtained by segmenting full videos into subareas, including 296 real videos and 112 synthetic videos. Besides, to simplify the annotation, each video is sampled by 1 fps. Finally, TMS contains 9,296 images. The number of real images and synthetic images are 7,336 and 1,960, respectively.
\begin{figure}[tp]
	\centering
	\includegraphics[width=0.48\textwidth]{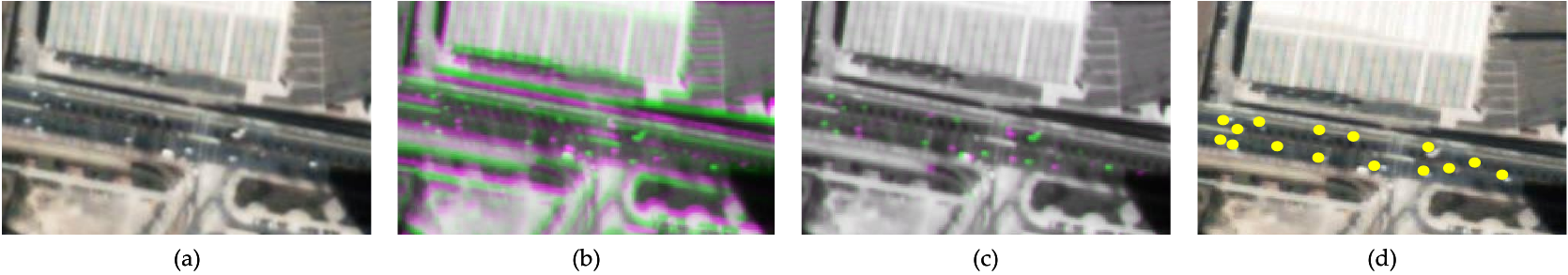}
	\caption{The annotation process of each real image in TMS. (a) is the original image. (b) shows the difference between consecutive frames. (c) is the difference of registered consecutive frames. (d) displays the final annotation of vehicles. }\label{Fig3}
\end{figure}

\begin{table*}[htp]
	\footnotesize
	\centering
	\caption{Statistics of reported datasets of traffic monitoring from satellite. }
	\renewcommand\arraystretch{1}
	\label{Table1}
	
	\begin{tabularx}{0.9\textwidth}{p{1.6cm}<{\centering}|X<{\centering}|p{2.6cm}<{\centering}|X<{\centering}|p{0.6cm}<{\centering}|p{0.6cm}<{\centering}|p{0.8cm}<{\centering}|p{0.6cm}<{\centering}|p{0.6cm}<{\centering}|p{0.6cm}<{\centering}|X<{\centering}}
		\hline
		
		\hline
		\multirow{2}{*}{Datasets} &\multirow{2}{*}{\#Videos} &\multirow{2}{*}{\#Resolution} &\multirow{2}{*}{\#Images}&\multicolumn{3}{c|}{\#Vehicles} &\multicolumn{3}{c|}{Tasks}&\multirow{2}{*}{Availability}\\
		\cline{5-7}	\cline{8-10}	
		&&&& Min&Max & Total &TOD&VEC&TDE\\ 
		\hline
		
		\hline

		SHDV \cite{zhang2018satellite} & 1 &400*400 &700 &--&--&--  &\Checkmark&\XSolidBrush  &\XSolidBrush& No\\\hline	
		Valencia \cite{ao2019needles} & 3 &500*500  &168 &7&41&3,211  &\Checkmark&\XSolidBrush  &\XSolidBrush& Yes\\\hline
		Las Vagas \cite{zhang2020error} & 2 &400*400, 600*400 &1,400 &27&86&80,047  &\Checkmark&\XSolidBrush  &\XSolidBrush& Yes\\\hline
		Jilin-1 \cite{zhang2021moving} & 3 &(400$ \sim $700)*(400$ \sim $600)  &900 &--&--&--  &\Checkmark&\XSolidBrush  &\XSolidBrush& No\\\hline
		SkySat \cite{zhang2021moving} & 6 &(400$ \sim $600)*(400$ \sim $600)  &3,500 &--&--&--  &\Checkmark&\XSolidBrush  &\Checkmark& Partially\\
		\hline
		
		\hline
		TMS  & 408 &512*512, 540*480 &9,296 &0&101&128,801 &\Checkmark&\Checkmark  &\Checkmark& Yes\\
		\hline
		
		\hline
	\end{tabularx}
\end{table*}
\begin{figure}[tp]
	\centering
	\includegraphics[width=0.49\textwidth]{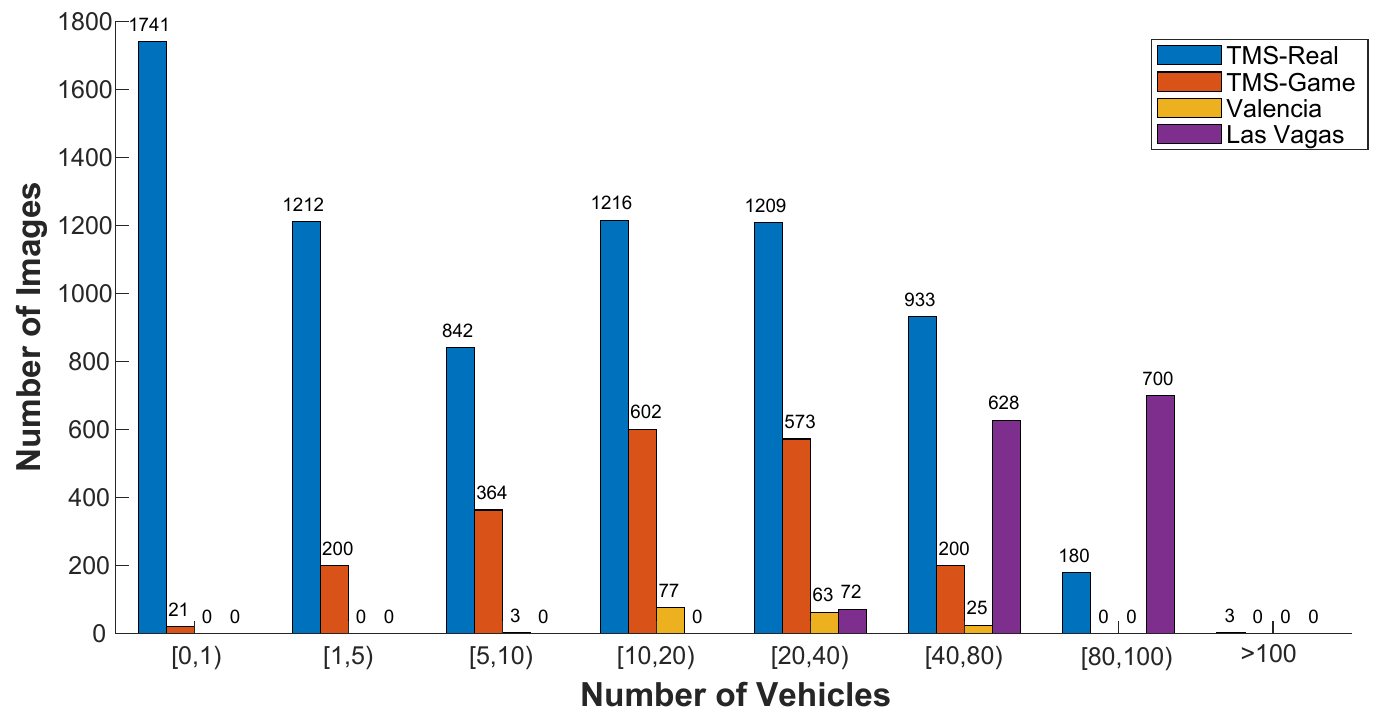}
	\caption{Image distribution of different datasets over the number of vehicles. Note that TMS-Real and TMS-Game represent the real and synthetic parts of TMS, respectively.}\label{Fig4}
\end{figure}

The annotation of satellite videos is quite arduous. It is because the vehicles are tiny and lack of appearance features, so that they can hardly be recognized from background. To address this problem, a motion based annotation method is conducted in our work. Specifically, the motion of vehicles is larger than the surrounding background. In this case, the vehicles can be distinguished from background by comparing the difference between two consecutive frames. To amplify the difference, two frames with the temporal interval of one second are compared, so that the difference can be recognized by human eyes. As depicted in Fig. \ref{Fig3}, the differences are highlighted by colors. The green color means the pixel value of current frame at this region is higher than the next frame, which indicates the location of vehicles. In contrast, the red color indicates the vehicle locations of next frames. However, the motions from vehicles and background are intertwined. It makes the annotation in a complete daze, as shown in Fig. \ref{Fig3} (b). 

To further reduce the interference caused by background motion, the translation of background is removed by the intensity-based image registration method\footnote{\url{https://www.mathworks.com/help/images/ref/imregister.html\#description}}. In this case, the annotator can localize the vehicles effortlessly. Each vehicle is annotated with a point which indicates the location. Practically, the annotation tool is developed based on Matlab R2019b. 

In synthetic videos, the locations of vehicles are obtained automatically by transforming the map coordinate into the screen coordinate. The annotation tool is designed based on the game plugin developer, Script Hook V \cite{ScriptHOOK}. It can save much labor force in the annotation process. Besides, it can generate lots of automatically annotated synthetic videos, in order to augment the training data and boost the performance on real satellite videos. It provides an efficient way to remedy the lack of real data in VHR satellite videos.

We want to emphasize that, different from the previously released Valencia dataset \cite{ao2019needles}, the IDs of vehicles in different frames are not provided in the proposed TMS dataset\footnote{Practically, our plugin is able to identify the vehicles in different frames.}. It is for the following reasons: 1)  Valencia is a partially annotated dataset, where the annotated regions are selected manually. In contrast, TMS is a fully annotated dataset. The vehicles are very dense and lack of appearance features. It can hardly be identified by annotators. 2) The VHR videos are captured by satellites in low earth orbit. The view of satellite is non-stationary, so that it can only gaze the city at a quite short time, which makes it infeasible to track vehicles effectively. In this case, the vehicle tracking task is not developed in this paper.

Overall, in our work, 14 annotators are employed in the annotation process, and more than 500 working hours are spent. Furthermore, each annotator also plays the role of checker, in order to guarantee the correctness of annotations for each image.
\begin{figure}[tp]
	\centering
	\subfigure[]{
		\label{Fig7(a)}
		\includegraphics[width=0.20\textwidth]{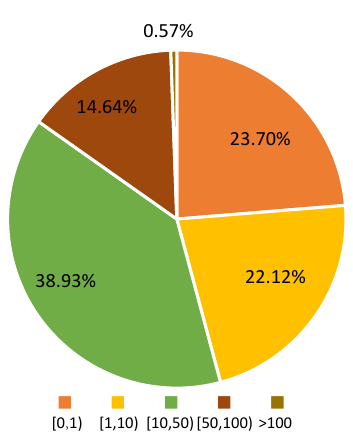}}
	\hspace{.2in} 
	\subfigure[]{
		\label{Fig7(b)}
		\includegraphics[width=0.20\textwidth]{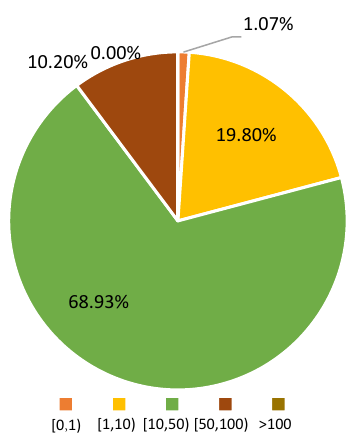}}
	\caption{Vehicle density distribution of TMS-real (a) and TMS-game (b).}\label{Fig7}
\end{figure}

\subsection{Data Statistics}

The TMS dataset is constructed from 26 real and synthetic satellite videos, where 408 video clips and 9,296 images are obtained. In each image, the locations of vehicles are annotated, and the number of vehicles varies from 0 to 101. In total, 128,801 vehicles are annotated.  

Table \ref{Table1} presents the statics of existing datasets. At present, TMS is the largest dataset for traffic monitoring from satellite. It is also the only one  publicly available large-scale dataset in this task. Compared to existing datasets, it has the following advantages:

1) The real part of TMS is created based on 12 fully annotated satellite videos, each covers an area of several square kilometers, and 7,336 annotated images are obtained. However, existing datasets only contain one or two partially annotated videos. With the large-scale TMS dataset, the deep learning approaches are applicable to the tasks of traffic monitoring from satellite. The evaluation can also be conducted effectively.

2) TMS is the first to utilize synthetic videos to augment the dataset. It can address the problem of lacking of publicly available satellite videos. Synthetic videos can be annotated automatically, which can reduce the labor costs significantly. By utilizing synthetic videos, TMS pioneers researchers a new perspective to develop approaches and boost the performance.

3) In TMS, the videos are captured over different cities all around the world as well as the virtual city. The scenes include downtown, suburbs, harbor, airport, \emph{etc.}. The vehicles in each image vary in a large range from 0 to 101. However, existing datasets only contain one or two scenes with dense traffic, which cannot meet the demand of traffic monitoring in different places and situations. In this case, TMS can improve the generality of developed approaches.

Furthermore, Fig. \ref{Fig4} and \ref{Fig7} plot the distribution of samples over the number of vehicles and traffic density. It can be observed that TMS holds the largest range over the number of vehicles among different datasets, \emph{i.e.,} from 0 to 100+. Besides, only TMS contains negative samples in which the number of vehicles is zero. The other two datasets, Valencia and Las Vagas, lack of null and sparse traffic scenes, since most of the samples are with the vehicles over 10. Furthermore, we can see from Fig. \ref{Fig7} that both the real part and synthetic part of TMS cover the null, sparse, normal and dense traffic scenes. It shows the superiority of TMS compared with existing publicly available datasets. In conclusion, TMS is more applicable to the real scenarios of traffic monitoring from satellite.

\subsection{Applicable Tasks}

To monitor traffic from satellite effectively, three tasks are developed based on the TMS dataset. They are Tiny Object Detection (TOD), VEhicle Counting (VEC) and Traffic Density Estimation (TDE). For the TMS dataset, we randomly allocated 50$\%$ as the training set, 25$\%$ as the valiadation set, and the rest for testing, adhering to this criterion across all tasks. Note that the dataset is allocated in video clip-wise (\emph{i.e.}, 408 clips totally) while not image wise, since two temporally adjacent images are similar and may result in training information leakage.

The target of TOD is to localize the vehicles in each image. It can be conducted by two strategies. One is the single-image detection strategy, where the objects are detected by applying detectors to single image. The other is motion-based detection strategy. The detectors operate on the video. The motion among consecutive frames is utilized to distinguish objects from background. Following existing detection protocols, the annotated points are modified to bounding boxes with the height and width of six pixels. The size of bounding boxes is large enough to cover most vehicles. In the evaluation, Precision, Recall and F-score are utilized as the metrics.%, which are defined as follows,

The task of VEC is to count the number of vehicles in the screen. It is a basic factor to estimate the 
traffic congestion. Different from TOD, VEC focuses on the global traffic situation, which is not necessary to localize each vehicle in the image. In this paper, the Mean Average Error (MAE) and Mean Square Error (MSE) are adopted as the evaluate metric, which can measure the difference of the esimated number and annotated number.

TDE aims to estimate the traffic density map in each image. It provides a vivid visualization of traffic congestion. In practice, to generate the ground truth of density maps, the annotated location marks are blurred by the Gaussian kernel, where the kernel size is fixed as 29, and \(\sigma = 4\). Two popular metrics are adopted in TDE, \emph{i.e.,} Structural Similarity in Image (SSIM) and Peak Signal-to-Noise Ratio (PSNR). They measure the performance of density map estimation.

\section{Experiments}\label{section4}

\begin{figure*}[tp]
	\centering
	\includegraphics[width=1.0\textwidth]{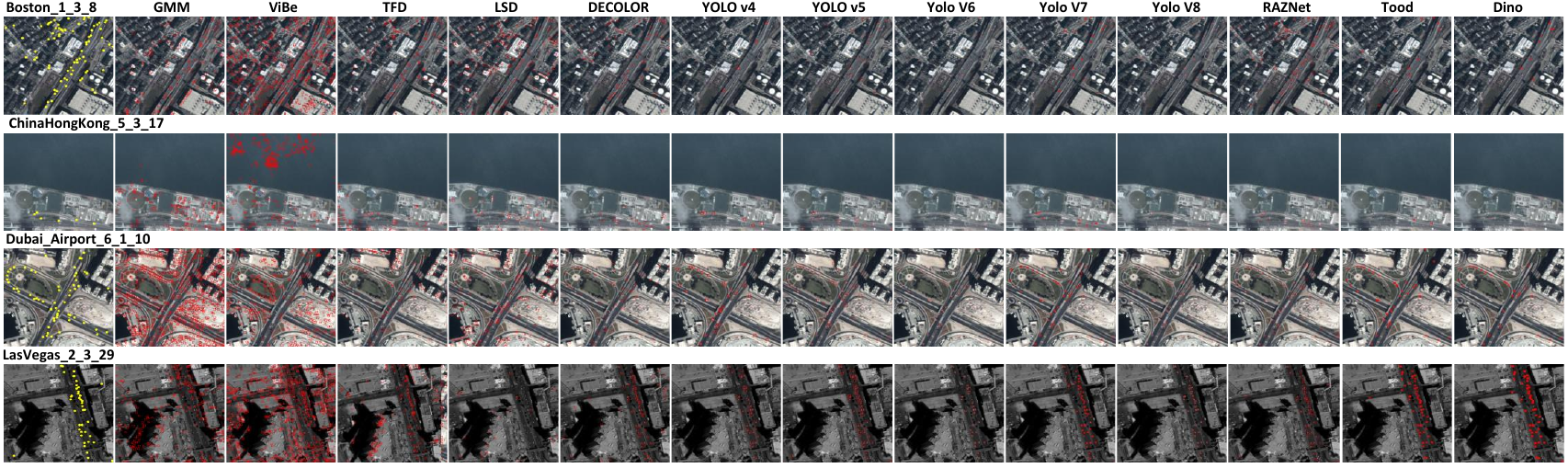}
	\caption{Visualization exemplar results of different approaches on the TOD task across the TMS dataset. The first column denotes the annotated detection results, and the other columns represent the final detection results of different methods. }\label{Fig8}
\end{figure*}
\begin{table*}[htp]
	\centering
	\caption{TABLE II TOD RESULTS OF DIFFERENT APPROACHES ON TMS. THE BEST RESULTS ARE MARKED IN BOLD.}\label{Table2}
	\renewcommand\arraystretch{1}
	
% 	\begin{tabularx} {0.9\textwidth}{p{3cm} 
% <{\centering}|X<{\centering}X<{\centering}X<{\centering}|X<{\centering}X<{\centering}X<{\centering}|X<{\centering}X<{\centering}X<{\centering}}
    \begin{tabular}{c|c|c|c|c|c|c|c|c|c}
		\hline		
		\hline
		Setting &\multicolumn{3}{c|}{Real}&\multicolumn{3}{c|}{Game}&\multicolumn{3}{c}{Augmented}\\\hline
		
		\hline
		Metrics &$Precision$$\uparrow$ & $Recall$$\uparrow$  &$F_{score}$$\uparrow$ &$Precision$$\uparrow$ & $Recall$$\uparrow$ &$F_{score}$$\uparrow$ &$Precision$$\uparrow$ & $Recall$$\uparrow$ &$F_{score}$$\uparrow$ \\\hline
		R-PCA \cite{candes2011robust} &6.67 &0.59 &0.97&1.43 &4.64 &2.03&$--$ &$--$ &$--$  \\\hline
		GMM \cite{rasmussen1999infinite} &23.60 &3.07&5.19&15.09 & 4.18 &6.15 &$--$ &$--$&$--$ \\\hline
		ViBe \cite{barnich2010vibe} &54.63 &9.51&14.92&30.68 & 17.56 &21.49 &$--$ &$--$&$--$  \\\hline
		TFD \cite{du2017object} &\textbf{69.33} &9.28&15.63&7.96 &2.69 &3.89 &$--$ &$--$&$--$ \\\hline
		LSD \cite{DBLP:journals/tip/LiuZYQ15}&46.15 &13.61&19.61&38.00 &29.76 &32.98 &$--$ &$--$&$--$ \\\hline
		DECOLOR \cite{zhou2012moving} &52.48 &\textbf{56.65}&\textbf{52.67}&32.21 & 46.01 &37.23 &$--$ &$--$&$--$ \\
		\hline
		
		\hline
		Faster RCNN \cite{DBLP:conf/nips/RenHGS15} &2.56 &3.92&3.00&1.68 &7.13 &3.00 &1.91 &3.39&2.00\\\hline
		SSD \cite{DBLP:conf/eccv/LiuAESRFB16}&3.20 &3.55&1.00&0.25 &3.41 &0.63 &0.32 &3.58&1.00\\\hline
		YOLO v4 \cite{bochkovskiy2020yolov4}&35.99 &14.75&22.00&39.41 &15.71 &22.00 &45.21 &23.52&31.00\\ \hline
  
  YOLO v5 & 39.8&27.1 & 32.2& 24.2&16.3&26.1 & 40.6&28.5&33.5\\ \hline

    YOLO v6 \cite{yolov6}&35.11 & 31.3 &36.43 &37.63 & 17.85 & 31.06  & 43.5&35.2 &37.6\\ \hline
  YOLO v7 \cite{yolov7}&28.03 &31.1  &29.45 &22.4 & 17.2&19.47 & 31.6&23.5  &33.2\\  \hline
		YOLO v8 \cite{yolov8}&35.61 & 32.71 & 39.81 &\textbf{40.13} & 19.63& 25.82 &47.1 &28.4&35.9\\ \hline
  
  Dino\cite{dino}&5.4 & 45.9 &9.7 &6.3 & 40.9& 10.9& 6.92 & 49.1&12.3 \\ \hline
		Tood \cite{TOOD}&14.0 &40.3  & 20.7& 8.8& 33.1&13.9 &19.3 &46.5 &22.8\\

		\hline
		
		\hline
		SCAL\_Net\cite{wang2021dense}&0.34 &6.92&2.17&0.33 & 0.19 &0.24 &0.37 &5.47&2.16\\  \hline
		RAZNet \cite{liu2019recurrent}&42.77 &53.24&47.53&28.05 &\textbf{59.04} &\textbf{38.03} &\textbf{51.31} &\textbf{62.12}&\textbf{56.78}\\ \hline
         \hline	
        \end{tabular}%
	% \end{tabularx}	
\end{table*}
\subsection{Experiments on Tiny Object Detection}
In the TOD task, three kinds of approaches are evaluated, \emph{i.e.,} background subtraction, deep object detector, deep object localizer. They are described in the following subsections. 

\subsubsection{Background Subtraction}
\textbf{R-PCA}: Robust Principle Component Analysis \cite{candes2011robust}. It separates the background and foreground into a low-rank matrix and a sparse matrix, respectively. It is optimized jointly by the principle component pursuit and fast low-rank approximation. The foreground pixels are obtained by morphological segmentation.

\noindent\textbf{GMM}: Gaussian Mixture Model \cite{rasmussen1999infinite}. It is developed with the assumption that the each pixel follows the Gaussian distribution temporally. It models the background with several Gaussian distributions. One pixel is identified as the background point once it belongs to one of distributions, and the background is updated iteratively.

\noindent\textbf{ViBe}: Visual Background extractor \cite{barnich2010vibe}. It models each background pixel with several neighbors. In this case, the background can be initialized in single frame. With the assumption that stochastic models can simulate the uncertainty of pixel change, the background is updated with one random neighbor if one point is identified as the background.

\noindent\textbf{TFD}: Three Frame Difference \cite{du2017object}. It subtracts the background by detecting the changes among frame sequence. Three consecutive frames are utilized to compute the changes pairwisely, which can effectively reduce the interference caused by occlusions and noise. 

\noindent\textbf{LSD}: Low-rank and Structured sparse Decomposition \cite{DBLP:journals/tip/LiuZYQ15}. It decomposes the foreground and background with the regularization of low-rank norm and structured sparse norm. 

\noindent\textbf{DECOLOR}: DEtecting Contiguous Outliers in the LOw-rank Representation \cite{zhou2012moving}. It represents the video frames as a low-rank matrix with the assumption that the background are linearly correlated, and objects are detected as outliers.

\subsubsection{Deep Object Detector}
\textbf{Faster RCNN}: Faster Region-CNN \cite{DBLP:conf/nips/RenHGS15}. It is a popular anchor-based deep object detector. It firstly utilizes the Region Proposal Network (RPN) to generate region proposals, and then apply the Region-CNN to regress the coordinate of detected objects.

\noindent\textbf{SSD}: Single Shot multibox Detector\cite{DBLP:conf/eccv/LiuAESRFB16}. It is a one-stage object detector that localizes and recognizes objects jointly. It is much efficient than Faster RCNN. In this paper, VGG-16 is adopted as the backbone of SSD. 

\noindent\textbf{YOLO series}: The fourth, fifth, sixth, seventh, and eighth version of You Only Look Once \cite{bochkovskiy2020yolov4}. YOLO is an anchor-free deep object detector. It segments the whole image into grid cells, and regresses the localization directly. YOLO v4~\cite{bochkovskiy2020yolov4},  YOLO v5 \footnote{\url{https://github.com/ultralytics/yolov5}}, YOLO v6~\cite{yolov6}, YOLO v7~\cite{yolov7}, and YOLO v8~\cite{yolov8} are quite similar. They both utilize techniques, such as cropping, rotation, flipping, mosaic, \emph{etc.}, to augment the training data.

    \noindent\textbf{Tood}: Task-aligned one-stage object detection~\cite{TOOD}. It is a one-stage object detector ~\cite{TOOD}, which explicitly aligns object classification and localization in a learning-based manner. It utilizes ResNet as the backbone and follows an overall pipeline of backbone-FPN-head.

\noindent\textbf{Dino}: DETR with Improved deNoising anchOr boxes~\cite{dino}. It is a robust end-to-end approach for improved denoising anchor boxes in object detection, and Transformer is adopted as the backbone.

\subsubsection{Deep Object Localizer}
\textbf{RAZNet}: Recurrent Attentive Zooming Network \cite{liu2019recurrent}. It is originally proposed for crowd localization. In this paper, it is utilized to localize the vehicles just with the point annotations. It is good at detecting tiny objects by operating recursively on small image regions and zooming them into high-resolution.

\noindent\textbf{SCAL\_Net}: A Simple yet effective Counting And Localization
Network \cite{wang2021dense}. It proposes a joint framework for vehicle counting and localization, which tackles them as a pixel-wise dense prediction problem.
\begin{figure*}[tp]
	\centering
	\includegraphics[width=1.0\textwidth]{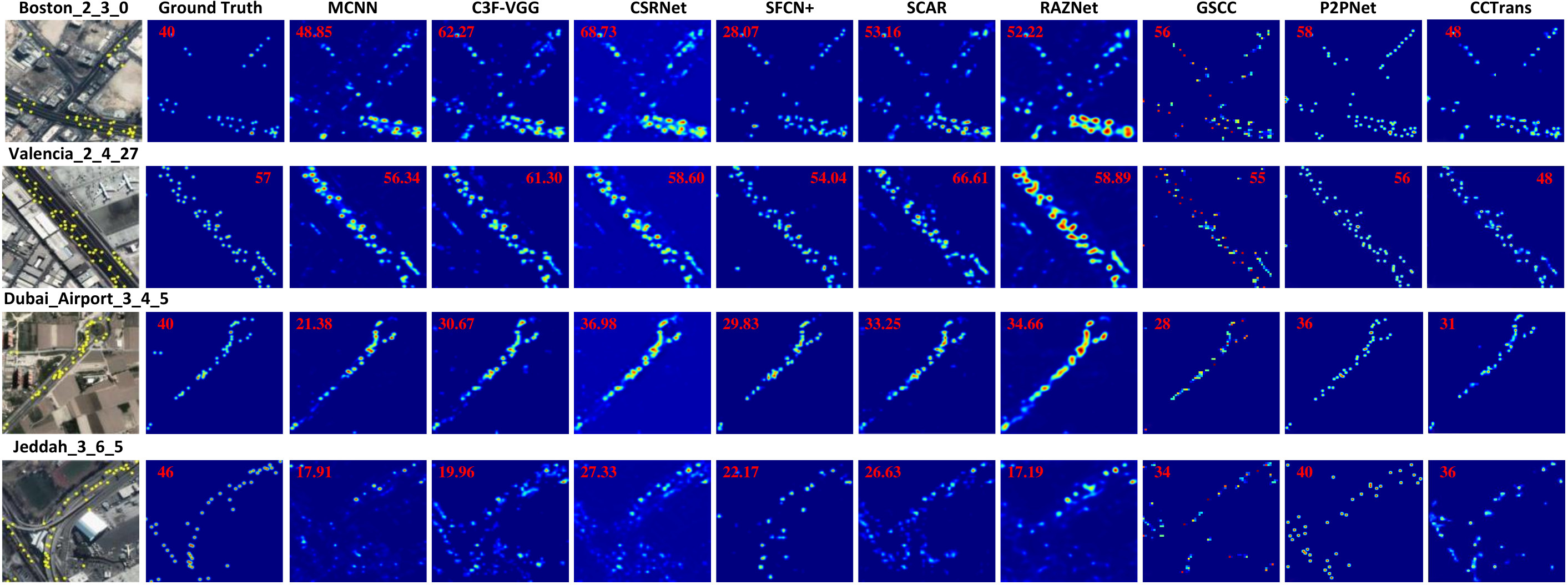}
	\caption{Visualization of the crowd density maps of different methods on the VEC and TDE tasks across the TMS dataset. The first and second columns denote the annotated density maps, and the other columns represent the final density estimation results of different methods. Note that the heat of each pixel indicates the vehicle density.
The red numbers denote the annotated and predicted counting results.}\label{Fig9}
\end{figure*}
\begin{table*}[htp]
	\centering
	\caption{VEC AND TDE RESULTS OF DIFFERENT APPROACHES ON TMS. THE BEST RESULTS ARE MARKED IN BOLD.}\label{Table3}
	\renewcommand\arraystretch{1}
	
	\begin{tabularx}{0.9\textwidth}{p{3cm}<{\centering}|X<{\centering}X<{\centering}|X<{\centering}X<{\centering}|X<{\centering}X<{\centering}}
		\hline
		
		\hline
		Setting &\multicolumn{2}{c|}{Real}&\multicolumn{2}{c|}{Game}&\multicolumn{2}{c}{Augmented}\\
		\hline
		
		Metrics &MAE/MSE $\downarrow$ & PSNR/SSIM $\uparrow$ &MAE/MSE$\downarrow$  & PSNR/SSIM$\uparrow$ &MAE/MSE $\downarrow$ &PSNR/SSIM $\uparrow$ \\\hline
		
		\hline
		MCNN \cite{zhang2016single} &7.15/113.83 &29.02/0.90 &7.28/107.37 &30.89/0.93
		&6.92/106.43 &29.33/0.90  \\\hline
		C3F-VGG \cite{gao2019c} &6.99/97.75&29.09/0.90&10.23/189.95&30.37/\textbf{0.97} & 5.98/83.25	&29.31/0.93\\\hline
		CSRNet \cite{li2018csrnet} &6.75/139.18&29.31/0.89&10.46/195.96&31.03/0.94&6.02/77.83&29.36/0.94  \\\hline
		SFCN+ \cite{wang2019learning} &\textbf{4.75/70.62}&28.95/\textbf{0.97}&11.28/234.42&30.65/\textbf{0.97} & 12.34/287.61 &28.66/0.79  \\\hline
		SCAR \cite{gao2019scar} &6.70/146.64 &29.70/0.96&8.01/125.58&30.58/0.94&
		5.79/105.49 &29.71/0.96\\\hline
		RAZNet \cite{liu2019recurrent}&7.17/102.47 &\textbf{30.17}/0.94&7.85/130.33&29.79/\textbf{0.97} &6.74/96.57  &\textbf{31.28/0.97}\\\hline
		SCAL\_Net\cite{wang2021dense}&13.02/354.57&8.74/0.56&10.87/182.47&17.57/0.78 & 12.97/329.33 &9.23/0.67 \\\hline

  GSCC\cite{GSCC} & 5.19/78.61   & 28.22/0.87 & 5.28/97.39 & 29.55/0.89  & \textbf{ 4.89}/78.05 & 29.25/0.91 \\\hline

  P2PNet\cite{P2PnET} & 5.96/117.36  &29.87/0.91  & 5.38/105.1 & \textbf{31.42}/0.96 & 5.03/\textbf{74.29} & 30.76/0.95 \\\hline

  CCTrans\cite{tian2021cctrans} & 5.34/ 84.37  &29.57/0.92 &  \textbf{4.42/59.62}& 29.87/0.95  & 5.96/87.35 & 29.54/0.92  \\
		\hline
		
		\hline
	\end{tabularx}
	
\end{table*}

\subsubsection{Results and Discussions}

Table \ref{Table2} presents the results of different approaches on the task of tiny object detection. It can be observed that TOD is really a challenging task, since most of the classic and popular approaches perform worse than they are on traditional object detection tasks. For background subtraction approaches, R-PCA gets really poor performance. It is mainly because the foreground in R-PCA is obtained by morphological segmentation, while these vehicles are so tiny that their morphological features can hardly be recognized. Worse still, other background modeling approaches, including GMM, ViBe and LSD, also suffer from the extremely unbalanced distribution of foreground and background pixels. TFD distinguishes vehicles from background by the changes among three consecutive frames. However, as aforementioned, the motion in satellite videos are quire complex. DECOLOR takes vehicles as outliers with the assumption that the background are linearly correlated in the frame sequence, which performs the best in background subtraction approaches.

Deep learning based object detection has achieved great success in the computer vision field. However, TOD is an extremely challenging task, since the vehicles in satellite videos are too tiny to be captured by the detector. It can be observed that Faster RCNN and SSD get poor performance, even worse than traditional detectors. It is mainly because that the scales of vehicles only take a minor part of the whole image. The appearance of vehicles is vanished from feature maps after the encoding of several convolutional and pooling layers. To maintain the vehicle features, YOLO v4, v5, v6, v7, and v8 utilize a more powerful and deeper backbone network to extract richer feature representations. In this case, their performance surpasses that of Faster RCNN and SSD, as well as outperforms Dino and Tood, which utilize multi-scale feature interactions.

The results of two deep object localizers are also presented in Table \ref{Table2}. They detect vehicles from point annotations, which is quite different from object targeting approaches. We can see that SCAL\_Net performs poor. It is because the vehicles are tiny and much smaller than the instance in the crowd. RAZNet achieves much better results on the real part of TMS. It is mainly benefited from its zooming strategy, which is operated on small regions and zooming into high-resolution recursively. 

Fig. \ref{Fig8} presents the results of different approaches on the TOD task. Note that those approaches perform extremely poor in Table \ref{Table2} are not displayed. We can see that the results of GMM, ViBe and TFD contain lots of false positive samples. Others performs better but far from satisfactory. Overall, from the results in Table \ref{Table2} and Fig. \ref{Fig8}, we can simply conclude that TOD is a challenging task due to the tiny size of vehicles. As a result, most traditional approaches get much worse performance than they are on the typical object detection task. Besides, deep learning based detectors and localizers show less superiority as well. Fortunately, spatial zooming and feature fusion provide some meaningful insights for future researches.

\subsection{Experiments on Vehicle Counting and Traffic Density Estimation}
Vehicle counting and density estimation are two relevant tasks. They follow similar protocols in the literature. In this case, the experiments of VEC and TDE are conducted and discussed in this subsection together.
\subsubsection{Methods for Comparison and Evaluation}
\textbf{MCNN}: Multi-Column CNN \cite{zhang2016single}. By designing kernel in different sizes, a multi-column scheme is conducted, which can adapt to arbitrary crowd density and  perspective map of images.

\noindent\textbf{C3F-VGG}: A variant of VGG-16 \cite{gao2019c}. The first 10 convolutional layers of VGG-16 are adopted for representation learning, and another two regression layers are added to estimate the density map. 

\noindent\textbf{CSRNet}: Congested Scene Recognition Network \cite{li2018csrnet}. It also takes VGG-16 as the backbone, and designs a dilation module on the top.

\noindent\textbf{SFCN+}: Spatially Fully Convolutional Network \cite{wang2019learning}. It takes ResNet-101 \cite{he2016deep} as the backbone, and adds dilation convolution, spatial encoder and regression to predict the density map directly.  

\noindent\textbf{SCAR}: Spatial-Channel-wise Attention Regression \cite{gao2019scar}. It operates self-attention on the spatial axis and channel axis of feature map, which can capture the global contextual information. 

\noindent\textbf{GSCC}. Generalized losS function for Crowd
Counting~\cite{GSCC}. It learns density map representations through the optimization of the non-equilibrium problem, and introduces a generalized loss function for acquiring density maps used in crowd counting and localization.

\noindent\textbf{P2PNet}. Point to Point Network~\cite{P2PnET}. It presents a purely point-based framework, allowing for a more precise and seamless integration of individual localization with crowd counting.

\noindent\textbf{CCTrans}. Crowd Counting with Transformer~\cite{tian2021cctrans} It leverages a pyramid visual transformer backbone to capture overarching crowd information, followed by the utilization of another pyramid visual transformer backbone to further encapsulate global crowd insights. 

\noindent\textbf{RAZNet} and \textbf{SCAL\_Net}.

\subsubsection{Results and Discussions}

Table \ref{Table3} presents the results of different approaches on vehicle counting and traffic density estimation. Ten recently proposed approaches are compared totally. They are all effective deep learning approaches adopted from the typical computer vision task, \emph{i.e.}, crowd counting. It can be observed that most of them perform comparably. Specifically, SFCN+ performs the best on the real part of TMS in vehicle counting task, which is a spatially fully convolutional network. RAZNet outperforms most of the others in traffic density estimation. It demonstrates the effectiveness of the zooming strategy in tiny vehicle perception once again. While CCTrans achieves the best results in terms of MAE and MSE on the game part of TMS, its performance on real part is subpar. 

Besides, most approaches perform better on the game part of TMS rather than the real part, although much more training samples are provided in the real part. It is mainly because the number of vehicles varies largely in the satellite videos. Worse still, the satellite videos contain lots of noise. In contrast, the game videos are much simple. Furthermore, we can see that the performance of most approaches is boosted by conducting the augmentation strategy. It verifies the feasibility of utilizing synthesized data to make up for the lack of real satellite videos.

Fig. \ref{Fig9} presents the results of different approaches on the VEC and TDE task. Note that SCAL\_Net performs extremely bad, so its results are not displayed. It can be observed that most of these deep learning based approaches perform well to localize the vehicles and estimate the number as well as the density. It provides an effective way to monitor the traffic from satellite.

\section{Conclusion and Outlook}\label{section5}
The VHR videos captured by commercial satellites enable traffic monitoring from satellite. To boost the development, we build a large-scale dataset, \emph{i.e.}, TMS. It is composed of 12 satellite videos and 14 synthetic videos. 9296 images are obtained and fully annotated with 128,801 vehicles.  It enables the application of deep learning approaches in this field and provides a new perspective to augment data from GTA-V. Three tasks are developed based on TMS, including tiny object detection, vehicle counting and density estimation. Several classic and popular approaches are tested. The results have demonstrated the challenges and superiority of TMS. The TMS dataset, annotation tools and developed tasks provide insights for traffic monitoring from satellite.

Based on the results derived from both quantitative and qualitative analysis, we have discerned captivating patterns and unveiled new challenges that demand consideration within the task of vehicle perception from satellite.

1) \emph{How to compensate for the performance variance of different regions?} The satellite videos captured over different cities are quite different, which may cause interference to the generality of trained models and their performance vary largely. To address this problem, researchers can place particular emphasis on addressing the challenge of vehicle perception from satellite videos through domain adaptation. Furthermore, they can delve deeper into the extraction of more effective domain-invariant features from the interplay between synthetic data and real-world data.

2) \emph{How to introduce visual foundation models to this task?} Recently, several visual foundation models are proposed, such as SAM \cite{kirillov2023segment}, Dino \cite{dino}, GPT-4V \footnote{\url{https://openai.com/research/gpt-4v-system-card}}. They exceed the performance of visual perception and outperform traditional and deep learning methods significantly, which lead the datacentric research in computer vision. Researchers are encouraged to employ or tune these foundation models to the tasks of vehicle perception from satellite.

3)  \emph{How to effectively address the satellite's view displacement issue?} As mentioned in the paper, VHR videos are captured by low Earth orbit satellites. The satellite's perspective is non-stationary, allowing only a brief gaze upon the city. This renders effective vehicle tracking unfeasible. In this context, the paper did not embark on vehicle tracking tasks. Future researchers may employ techniques such as image stitching and image alignment to integrate data from multiple orbits into a coherent and real-time traffic monitoring system, enabling continuous surveillance of traffic within satellite videos.

\bibliographystyle{IEEEtran}
\bibliography{bare_adv}

\end{document}